# MULTI-SCALE SPATIALLY WEIGHTED LOCAL HISTOGRAMS IN O(1)

*Mahdieh Poostchi[1], Ali Shafiekhani[1], Kannappan Palaniappan[1], Guna Seetharaman[2]*

[1]Electrical Engineering and Computer Science Department
[1]University of Missouri-Columbia, MO, USA
[2]US Naval Research Laboratory, Washington D.C., USA

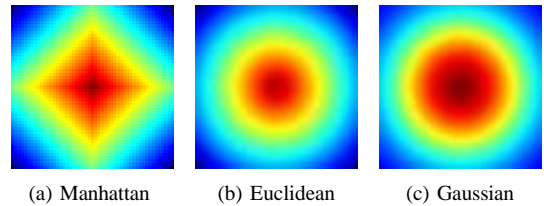

(a) Manhattan  (b) Euclidean  (c) Gaussian

**Fig. 1**. Illustration of linear and non-linear distance kernels.

## ABSTRACT

Weighting pixel contribution considering its location is a key feature in many fundamental image processing tasks including filtering, object modeling and distance matching. Several techniques have been proposed that incorporate Spatial information to increase the accuracy and boost the performance of detection, tracking and recognition systems at the cost of speed. But, it is still not clear how to efficiently extract weighted local histograms in constant time using integral histogram. This paper presents a novel algorithm to compute accurately multi-scale Spatially weighted local histograms in constant time using Weighted Integral Histogram (SWIH) for fast search. We applied our spatially weighted integral histogram approach for fast tracking and obtained more accurate and robust target localization result in comparison with using plain histogram.

## 1. INTRODUCTION

In many image processing applications, histograms are commonly used to characterize and analyze the region of interest within the image. Histogram-based features are space efficient, simple to compute, robust to translation and particularly invariant to orientation for color-based features. However, when computing a plain histogram, spatial information are missed which makes it sensitive to noise and occlusion. Several techniques are proposed to preserve spatial information including color Correlograms [1], Spatiogram [2], Multiresolution histogram [3], locality sensitive histogram [4, 5] and fragment-based approaches that exploit the spatial relationships between patches [6]. Spatially weighted histograms boost the performance of many image processing tasks at the expense of speed. In [7], Porikli generalized the concept of integral image and presented computationally very fast method to extract the plain histogram of any arbitrary region in constant time. Integral histogram provides an optimum and complete solution for the histogram-based search problem. Since then many novel approaches have been presented based on integral histogram to accelerate the performance of image processing tasks and incorporate the spatial information including filtering [8, 9, 10, 11], classification and recognition [3, 12], and detection and tracking [13, 14].

Despite all different techniques that have been proposed to adaptively weight the contribution of pixels when computing local histograms by considering their distance from center pixel, the problem of how accurately extract the spatially weighted histogram of any arbitrary region within an image in constant time using integral histogram is still unsolved. Fragtrack [6] proposes a discrete approximation scheme instead of the continuous kernel weighting approach to give higher weight to the contribution of inner rectangle compare to region margins for fast search.

In this paper we present a novel fast algorithm to accurately evaluate spatially weighted local histograms in O(1) time complexity using an extension of the integral histogram method (SWIH). The main idea is to (1) split local histogram kernel into multiple quadrants $q_i$ and decompose the spatial filter into independent weights $w_i$ subsequently, (2) for all $w_i$ compute candidate region weighted integral histogram $IH_{w_i}$, (3) then for every quadrant $q_i$ compute its weighted local histogram using the corresponding $IH_{w_i}$ and considering its translation from center pixel, (4) normalize the local histograms and (5) finally add them together to build the full kernel spatially weighted local histogram (Figure 2).

In the following section, we discuss the details of our new approach to extract the spatially weighted local histograms in constant time using integral histogram particularly for fast histogram-based similarity matching. Then, we evaluate the performance of SWIH and compare it with the brute-force implementation and approximation scheme in terms of computational complexity and accuracy.

## 2. SPATIALLY WEIGHTED LOCAL HISTOGRAMS

Weighting pixel contributions is a key feature in many fundamental image processing tasks including filtering, modeling

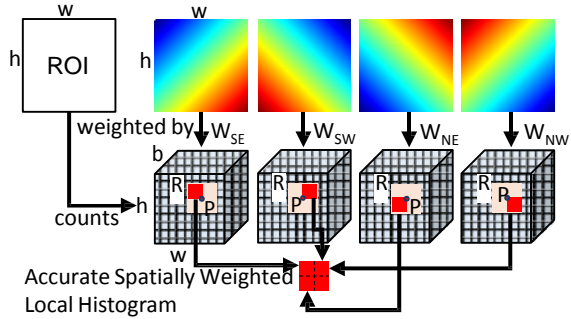

**Fig. 2**. Computational flow of accurate spatially weighted local histograms computation using weighted integral histogram for the region of interest (ROI) of size $w \times h$.

and matching to increase the accuracy of results in detection, tracking, recognition, etc..

The main idea is to assign lower weights to the pixels that most likely belong to background or occluding objects. One common technique is to define a weighting function that assigns weights to pixels with respect to their distance from target center (since undesirable pixels are usually considered around the region contours) including Manhattan, Euclidean, Gaussian or exponential weighting distance functions (Figure 1). Having such kernels enables us to adaptively weight the contribution of pixels and diminish the presence of background information when computing weighted local histograms. Figure 3 shows the accuracy of intensity feature likelihood maps based on sliding window histogram matching when using plain local histograms versus spatially weighted local histograms. As it can be seen, using spatially weighted local histograms generates more robust matching results. In the following sections, we describe the straightforward convolution-based approach, the discrete approximation scheme and our proposed novel, fast and accurate algorithm based on weighted integral histogram to compute spatially weighted local histograms for fast search.

*Brute-force Approach*: the computational complexity of the brute-force approach to compute the adaptively weighted local histograms at each candidate pixel location is linear in the kernel size and the number of candidate pixels. Assuming a search window of size $w \times h$ and a neighborhood of size $k \times k$ and $b$-dimensional histogram, the computational com-plexity of finding the best matched pixel location is $O(b \times k^2 \times w \times h)$, which makes the system far away from real-time performance particularly when it comes to large scale high resolution image analysis.

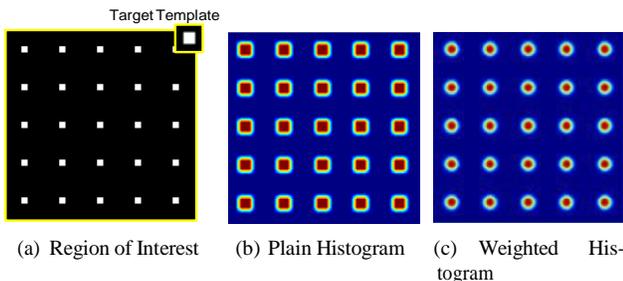

(a) Region of Interest    (b) Plain Histogram    (c) Weighted Histogram

**Fig. 3**. Performance evaluation of intensity feature likelihood maps using sliding window (b) plain versus (c) spatially weighted histogram distance matching.

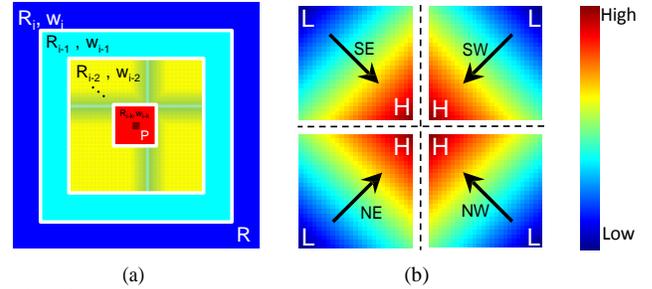

**Fig. 4**. (a) Wedding-Cake Approach: the discrete approximation scheme to obtain the spatially weighted local histogram for the candidate region considering inner-nested windows and using integral histogram ($w_i < w_{i-1} < ... < w_{i-k}$). (b) Tiling the kernel into four quadrants and decomposing the weights to accurately compute spatially weighted local histogram.

*Wedding-Cake Approach*: One solution to meet the demands of real-time implementation is to extract local histograms in constant time using integral histogram. However, as of our knowledge, there is still no solution to accurately and efficiently extract spatially weighted local histograms in $O(1)$ using integral histogram but the discrete approximate scheme presented in [6]. Frag-track proposed a simple approach to approximate the kernel function with different weights instead of pixel-level kernel weighting. Assuming that we want to calculate a spatially weighted local histogram in the rectangular region $R$ centered at point $P$ using integral histogram. Such counting can be approximated by considering several inner-nested windows $R_i$ at multiple scales around $P$ (Figure 4). The goal is to compute the counts of the rings between two adjacent windows $R_i$ and $R_{i-1}$ by subtracting their local histograms that are obtained in constant time using integral histogram. Then, rings histograms will be weighted appropriately with respect to their distance from $P$ and combined to provide an approximate spatially weighted local histogram on $R$. The accuracy of this approximation relies on the number of considered inner-nested windows.

In this paper, we present a new approach to compute spatially weighted local histograms that is more accurate than the wedding-cake method and takes constant time using integral histograms.

### 2.1. Multi-scale Spatially Weighted Local Histograms in Constant Time Using Integral Histogram (SWIH)

When using integral histogram, it is not clear how to weight pixel contributions when computing arbitrary rectangular region histogram in $O(1)$. We propose to address the pixel-level weighting problem by tiling the kernel into multiple quadrants as well as decomposing the weights (Figure 4(b)). In this section we describe our proposed algorithm in details when us-ing Manhattan distance function to adaptively weight pixel contributions (Figure 1(a)) for fast matching. Assuming that we want to weight the contribution of each pixel within region $R$ centered at $P_c = (x_c, y_c)$ by its Manhattan distance from $P$ when computing histogram of region $R$. Manhattan or city-block weighting function measures the sum of the absolute distance between two points along each axis. In our case, Manhattan distance of any arbitrary point $P_i = (x_i, y_i)$ withing region $R$ is

$$Dist_{Manhattan}(P_i, P_c) = |x_i - x_c| + |y_i - y_c| \quad (1)$$

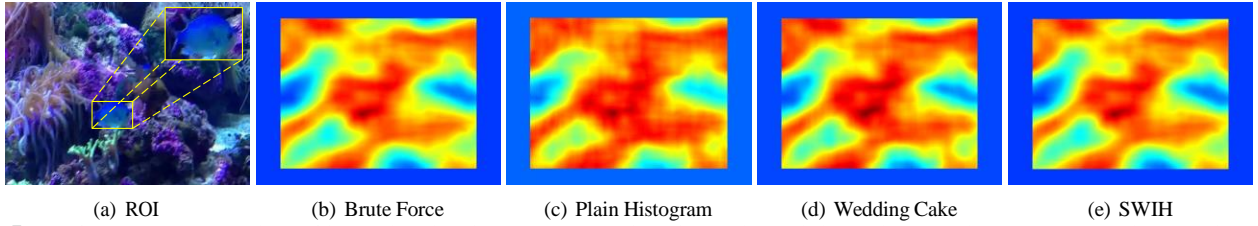

(a) ROI    (b) Brute Force    (c) Plain Histogram    (d) Wedding Cake    (e) SWIH

**Fig. 5**. Performance evaluation of intensity likelihood maps estimation using sliding window histogram matching. Weighting pixel contribution considering its location results in more accurate and robust target localization as shown in (b) and (e).

Since the filter is rectilinear and symmetric, we propose to decompose it into four independent weighting functions in the shape of four quadrants: TopLeft(TL), TopRight(TR), BottomLeft(BL) and BottomRight(BR) (Figure 4(b)). As it is shown, weights linearly increase from one corner to its diagonally opposite corner in each of the quadrants covering four directions: {SE, SW, NE, NW}. We extend these weights for the region of interest and compute four differently weighted integral histogram. For each direction, we consider two correlated images $f$ and $w_{dir}$ to compute the weighted integral histogram up to point $(x, y)$:

$$IH_{w_{dir}}(x, y, b_i) = \sum_{i \leq x, j \leq y} \delta(Q(f(i,j)) - b_i) w_{dir}(i,j) \quad (2)$$

$f$ contains image feature values, $Q$ is the quantization function that determines which bin to increase, $\delta$ is the impulse function and $w_{dir}$ is the pixel-wise weighing function that determine the value to increase at that bin. Having four differently weighted integral histogram, each of the quadrants spatially weighted local histogram will be computed in O(1) using its corresponding weighted integral histogram and considering its translation from the kernel center point. We will normalize the histograms and add them together to build the full region spatially weighted histogram. Figure 2 illustrates the flow of the computation. It is noteworthy to mention that due to weights rectilinear changes, their values are independent of the pixel location in the region of interests. This characteristic enables us to appropriately normalize the computed weighted local histogram and match it with the target spatially weighted histogram regardless of its location. This new method provides multi-scale accurate spatially weighted local histogram in constant time and can be utilized for other spatial weighting functions. It can be easily adapted to any fast computation of integral histogram on GPUs to accelerate the computation [15].

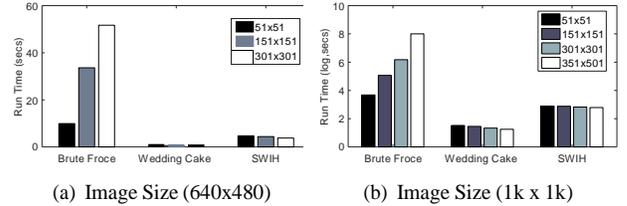

(a) Image Size (640x480)    (b) Image Size (1k x 1k)

**Fig. 6**. Performance comparison by increasing the local histogram kernel size. Computational time of integral histogram-based methods are invariant of kernel size.

## 3. PERFORMANCE EVALUATION AND EXPERIMENTAL RESULTS

In this section, we evaluate the performance of our approach and compare it with brute-force implementation and approximation scheme with respect to computational complexity and accuracy. Figure 5 evaluates the performance of the estimated intensity likelihood maps for a sample image from the VOT2016 data set [16] using sliding-window histogram matching. We compared the intensity likelihood map computed by the brute-force implementation with the matching results of the plain histogram, approximation scheme and our proposed accurate fast spatially weighted histogram. Background clutter is one of the main challenges in object detection systems relied on matching. We have selected an image that contains background clutter to make the matching process very challenging. It is shown that our proposed method not only provides exact results as the brute-force approach but is much faster. SWIH is 4.5 times faster than brute-force implementation for a candidate region of size $345 \times 460$ and sliding window of size $61 \times 91$ ( 5(a)). Figure 6 (a) and (b) shows the computational complexity of each of the discussed methods for standard image $640 \times 480$ as well as large scale image of size $1k \times 1k$ for different sliding window size from small scale to very large scale. It can be seen that the local histograms computational time of the brute-force implementation increases dramatically by enlarging the kernel size but is invariant of kernel size for the approximation scheme and SWIH.

## 4. SPATIALLY WEIGHTED INTEGRAL HISTOGRAM FOR FAST TRACKING (SWIFT)

Many of the generative region-based tracking algorithms rely on histograms for a fast and memory efficient appearance

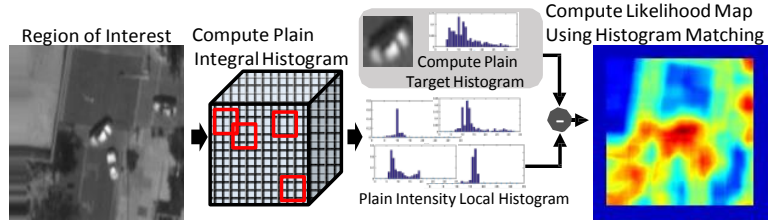

(a) Likelihood map estimation using sliding window plain histograms matching

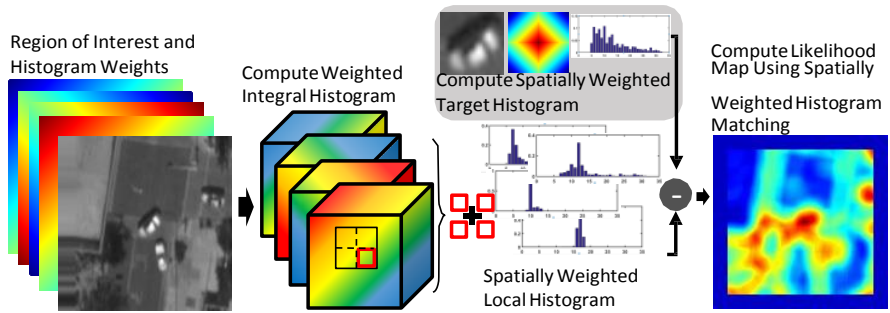

(b) Likelihood map estimation using sliding window spatially weighted histograms matching

**Fig. 7**. Performance evaluation of intensity likelihood map estimation using sliding window (a) plain histogram versus (b) our proposed accurate spatially weighted histograms matching approach for vehicle tracking in large scale aerial imagery.

modeling of target object as well as candidate region including Mean-Shift [17] and kernel-based [18, 19] approaches. However, since many of these trackers discard the spatial information when computing the conventional histogram, they rapidly lose the accuracy and converge to false targets. Therefore, many techniques have been presented to incorporate spatial information and enhance the tracker robustness which are either more computationally intensive and far away from real-time performance or a combination of different techniques to compensate the lack of spatial information [2, 20, 4, 13, 21]. Our proposed approach that retain spatial information as well as chromatic, shape or texture features presents a novel solution based on integral histogram that builds a robust and efficient histogram-based appearance model and provides accurate and fast histogram-based matching for visual tracking, detection and recognition purposes.

We integrated our proposed technique into our tracking system named *LoFT* [22, 23] to evaluate the performance of intensity histogram matching when using plain histogram versus spatially weighted histogram. *LoFT* is an appearance-based Likelihood of Features Tracking (LoFT) system, specialized for low resolution targets with large displacements caused by low frame rate sampling in Wide Area Motion Imagery(WAMI). Matching likelihood maps for individual features are computed using sliding window histogram similarity operators. The integral histogram method is used to accelerate computation of the sliding window histograms for a posteriori likelihood estimation [15]. Figure 7(a) describes the flow of the likelihood map computation using plain intensity histogram of target object and candidate regions. Similar to the results obtained for the synthetic image shown in Figure 3, using plain histograms results in less accurate localization of object. Hence, we applied our accurate spatially weighted integral histogram to estimate features likelihood maps instead of regular integral histogram for fast search. Figure 7(b) illustrates the computational flow compared to plain histogram and presents the more accurate target localization results when using spatially weighted histograms.

## 5. CONCLUSION

This paper present a novel fast algorithm to accurately evaluate spatially weighted local histograms in constant time using an extension of the integral histogram method (SWIH). We have shown that SWIH produces exact local histograms compared to brute-force approach and is much faster. Utilizing the integral histogram makes it to be fast, multi-scale and flexible to different weighting functions. This technique can be applied to fragment-based approaches to adaptively weight object patches considering their location. SWIH can be integrated into any detection or tracking system to provide an efficient exhaustive search and achieve more robust and accurate target localization.

## 6. ACKNOWLEDGMENT

This research was partially supported by the U.S. Air Force Research Laboratory under agreement AFRL FA875014-2-0072. The views and conclusions contained in this document are those of the authors and should not be interpreted as representing the official policies, either expressed or implied, of AFRL, NRL, or the U.S. Government.